# Offline Arabic Handwriting Recognition Using Artificial Neural Network

A.A Zaidan, B.B Zaidan, Hamid.A.Jalab, Hamdan.O.Alanazi and Rami Alnaqeib

**Abstract**— The ambition of a character recognition system is to transform a text document typed on paper into a digital format that can be manipulated by word processor software Unlike other languages, Arabic has unique features, while other language doesn't have, from this language these are seven or eight language such as ordo, jewie and Persian writing, Arabic has twenty eight letters, each of which can be linked in three different ways or separated depending on the case. The difficulty of the Arabic handwriting recognition is that, the accuracy of the character recognition which affects on the accuracy of the word recognition, in additional there is also two or three from for each character, the suggested solution by using artificial neural network can solve the problem and overcome the difficulty of Arabic handwriting recognition.

——————————   ◆   ——————————

## 1. INTRODUCTION

The goal of a character recognition system is to transform a text document typed on paper into a digital format that can be manipulated by word processor software [1]. The system is required to identify a given input data/character form by mapping it to a single character in a given character set [2]. This process can be quite involved since there are several valid forms that a character may take. This is largely due to the many fonts and styles (bold type, italic type, etc.….) that can be used. The motivation behind developing character recognition systems is inspired by its wide range of applications including archiving documents, automatic reading of checks, and number plate reading [3]. The field of handwriting recognition can be split into two different approaches. The first of these, on-line, deals with the recognition of handwriting captured by a tablet or similar touch-sensitive device, and uses the digitized trace of the pen to recognize the symbol [4].

- *A. A. Zaidan – PhD Candidate on the Department of Electrical & Computer Engineering , Faculty of Engineering , Multimedia University , Cyberjaya, Malaysia.*
- *B. B. Zaidan – PhD Candidate on the Department of Electrical & Computer Engineering / Faculty of Engineering, Multimedia University, Cyberjaya, Malaysia,*
- *Dr. Hamid.A.Jalab- Senior Lecturer, Department of Computer Science & Information Technology, University Malaya, Kuala Lumpur, Malaysia.*
- *Hamdan.O.Alanazi – Master Student, Department of Computer System & Technology, University Malaya, Kuala Lumpur, Malaysia, Email: Hamdan.*
- *Rami Alnaqeib – Master Student, Department of Information Technology, University Malaya, Kuala Lumpur, Malaysia.*

In this instance the recognizer will have access to the x and y coordinates as a function of time, and thus has temporal information about how the symbol was formed. The second approach concentrates on the recognition of handwriting in the form of an image, and is termed off-line [5]. In this instance only the completed character or word is available.

## 2. HANDWRITING RECOGNITION

optical character recognition, usually shortened to **OCR**, is the mechanical or electronic translation of images of handwritten, English, Arabic, or any other language has it owns characteristic in term of the latter shape, histogram, way of writing and number of letters, OCR mostly used AI tools to classify or give the dissuasion of translating the image characters into digital characters, two way are usually used for recognition, either through supervised or unsupervised system, the most well-known supervised methods is Neural Network , SVM, KNN, while Fuzzy logic is one of the unsupervised method. Supervised methods depend on train the system on a sample of data; in this case this operation would call training data, after the training has been done, the system will be ready to classify any object according the trained data. Data recognition in general fallow the below stages [6],[7],[8],[9],[10]:

A. In the training level:
   a. Segment character by hand to be ready for training
   b. Extract features
   c. Select features

B. Testing and recognition

    a. Input images include handwriting
    b. Test the system

Arabic handwriting is one of the most complex cases for recognition, unlike English writing; Arabic letters has many ways, no uppercase and lowercase, but the letter has three shapes, in the beginning of word, in middle, in the end. In additional there are much kind handwriting, an accurate feature should be extract and wide sample should be use for training to have an accurate result. Arabic handwriting recognition requires character segmentation as the position of each isolated character is unknown. Handwritten characters can be extracted from the rectangles and can be recognized by a character recognizer. Optical character recognition, usually abbreviated to OCR, is the mechanical or electronic translation of images of handwritten, typewritten or printed text (usually captured by a scanner) into machine-editable text. It is used to convert paper books and documents into electronic files, for instance, to computerize an old record-keeping system in an office, or to serve on a website such as Project Gutenberg. Arabic, one of the six official languages of the United Nations, is a Semitic language and the mother tongue of more than 300 million people. Written Arabic is standardized for most official communication in the Arab world. Its character set and similar sets are used by a much higher percentage of the world's population to write languages such as Arabic, Farsi, and Urdu. Recognition methods for current written Arabic can also be applied to ancient manuscripts [11].

Below are some notable features:
1. The Arabic alphabet contains 28 letters.
2. Words are written from right to left, numerals are written from left to right.
3. Most letters change their shape depending on whether they appear at the beginning, middle or end of a word, or on their own.
4. Letters that can be joined are always concatenated in both hand-written and printed Arabic. The only exceptions to this rule are crossword puzzles and diacritics in which the script is written vertically.

There are two main steps which must be done in order to achieve any recognition [12]:

## 2.1 Pre-processing

This step should cover all those functions carried out prior to feature extraction to produce a cleaned up version of the original image so that it can be used directly and efficiently by the feature extraction components of the Optical Characters Recognition (OCR) system [14].

## 2.2 Features Extraction

Features should contain information required to distinguish between classes, be insensitive to irrelevant variability in the input, and also be limited in number to permit efficient computation of discriminate functions and to limit the amount of training data required. This step involves measuring the features of the input character which are related to classification step. After features extraction, the character will be represented by a set of extracted features [15].

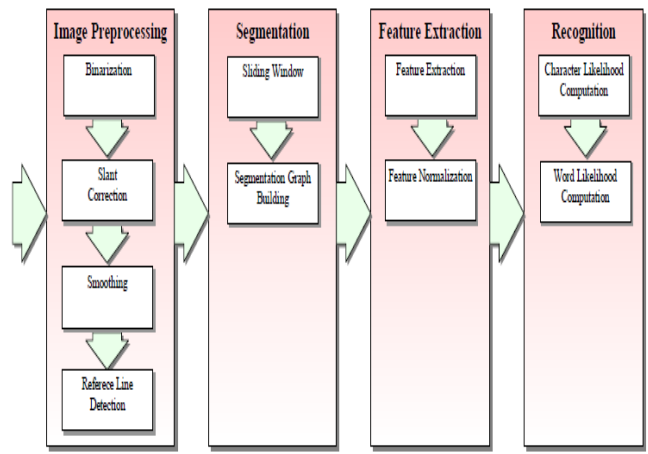

Fig.1 The General Processing of Handwriting Recognition

## 3. PROBLEM STATEMENT

Unlike other languages, Arabic has unique features, while other language doesn't have, from this language these are seven or eight language such as ordo, jewie and Persian writing, Arabic has twenty eight letters, each of which can be linked in three different ways or separated depending on the case [16].

Therefore, each character can have up to four different forms depending on its position. Furthermore, Arabic characters have different heights, which hamper noise detection.

## 4. METHODOLOGY

Machine Learning: This approach involves Machine Learning (ML) techniques, in which classification rules are automatically created, by using information from labeled (already-classified) texts. Artificial Neural Network (ANN) Neural network is an artificial intelligence technique simulating the way that human brain learns to recognize various patterns. ANNs were introduced by McCulloch and Pitts in 1943. ANN is trainable algorithms that can "learn" to solve complex problems from training data that consist of a set of pairs of inputs and desired outputs (targets). ANN has been applied successfully in many fields including speech recognition, image processing, and adaptive control. They can be trained to perform a specific task such as prediction, and classification. ANN consists of interconnected processing elements called neurons that work together to produce an output. There are large classes of algorithms which are applicable to (ANN) classification. In general, a neural network is a certain complex function that may be decomposed into smaller parts (neurons, processing units), and represented graphically as a network of these neurons. Segmentation and recognition might be used ANN to extract and classify the character from the image. The input of the ANN might be Data that been classified and some time features while the expect out-come is a classified character.

A part from the demo has been presented in the figure below.

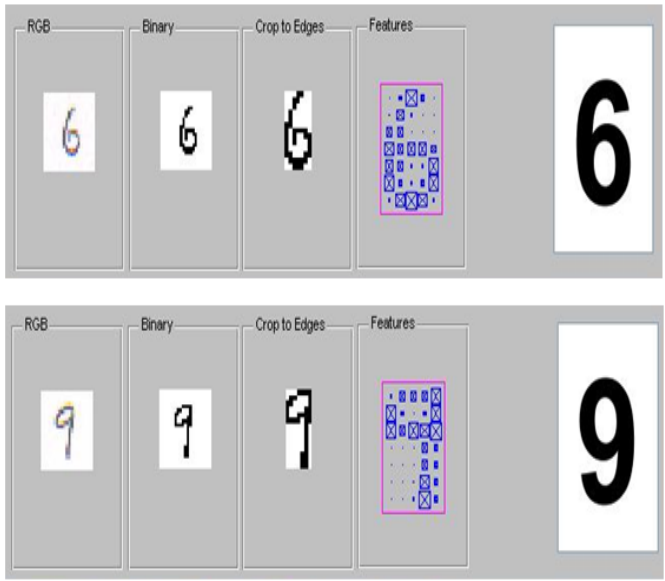

Fig. 2 Step by Step How the Demo Work

Below are the steps of the system flow:

1. In the first step we convert the selected part into RGB
2. Convert the RGB into binary image
3. Segments the non-black pixels
4. Extract the features
5. Input to the neural network
6. Recognition

## 5. CONCLUSION

In this paper artificial neural network successfully to solve the problem and overcome the difficulty of Arabic handwriting recognition and the ANN achieved 99.62% to the percentage for recognition using handwriting database.

## ACKNOWLEDGEMENT

The author would like to acknowledge all workers involved in this project and had given their support in more ways than one; the author would like to thank in advance his supervisor Dr. Hamid for her unlimited support, without her notes and suggestion this research would not be appear.

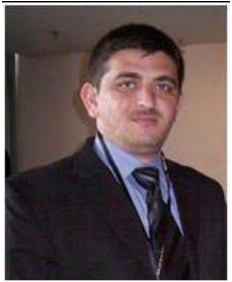

**Aos Alaa Zaidan**: He obtained his 1st Class Bachelor degree in Computer Engineering from university of Technology / Baghdad followed by master in data communication and computer network from University of Malaya. He led or member for many funded research projects and He has published more than 60 papers at various international and national conferences and journals, His interest area are Information security (Steganography and Digital watermarking), Network Security (Encryption Methods) , Image Processing (Skin Detector), Pattern Recognition , Machine Learning (Neural Network, Fuzzy Logic and Bayesian) Methods and Text Mining and Video Mining. .Currently, he is PhD Candidate on the Department of Electrical & Computer Engineering / Faculty of Engineering / Multimedia University / Cyberjaya, Malaysia. He is members IAENG, CSTA, WASET, and IACSIT. He is reviewer in the (IJSIS, IJCSN, IJCSE and IJCIIS).

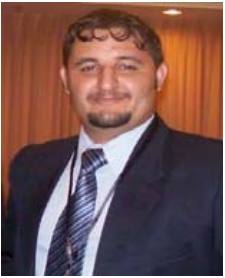

**Bilal Bahaa Zaidan:** He obtained his bachelor degree in Mathematics and Computer Application from Saddam University/Baghdad followed by master in data communication and computer network from University of Malaya. He led or member for many funded research projects and He has published more than 60 papers at various international and national conferences and journals, His interest area are Information security (Steganography and Digital watermarking), Network Security (Encryption Methods) , Image Processing (Skin Detector), Pattern Recognition , Machine Learning (Neural Network, Fuzzy Logic and Bayesian) Methods and Text Mining and Video Mining. .Currently, he is PhD Candidate on the Department of Electrical & Computer Engineering / Faculty of Engineering / Multimedia University / Cyberjaya, Malaysia. He is members IAENG, CSTA, WASET, and IACSIT. He is reviewer in the (IJSIS, IJCSN, IJCSE and IJCIIS).

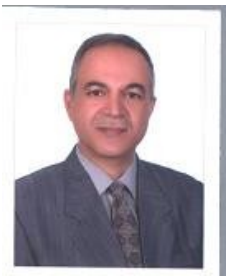

**Dr.Hamid.A.Jalab:** Received his B.Sc degree from University of Technology, Baghdad, Iraq. MSc & Ph.D degrees from Odessa Polytechnic National State University 1987 and 1991, respectively. Presently, Visiting Senior Lecturer of Computer System and Technology, Faculty of Computer Science and Information Technology, University of Malaya, Malaysia. various international and national conferences and journals, His interest area are Information security (Steganography and Digital watermarking), Network Security (Encryption Methods) , Image Processing (Skin Detector), Pattern Recognition , Machine Learning (Neural Network, Fuzzy Logic and Bayesian) Methods and Text Mining and Video Mining.

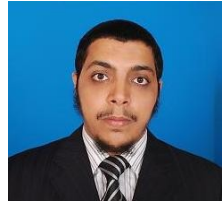

**Hamdan Al-Anazi:** has obtained his bachelor dgree from "King Suad University", Riyadh, Saudi Arabia. He worked as a lecturer at Health College in the Ministry of Health in Saudi Arabia, then he worked as a lecturer at King Saud University in the computer department. Currently he is Master candidate at faculty of Computer Science & Information Technology at University of Malaya in Kuala Lumpur, Malaysia. His research interest on Information Security, cryptography, steganography and digital watermarking, He has contributed to many papers some of them still under reviewer.

**Rami Alnaqeib** - he is master student in the Department of Information Technology / Faculty of Computer Science and Information Technology/University of Malaya / Kuala Lumpur/Malaysia, He has contribution for many papers at international conferences and journals.